\pdfoutput=1

\documentclass[11pt]{article}

\usepackage[final]{acl}

\usepackage{times}
\usepackage{latexsym}

\usepackage[T1]{fontenc}

\usepackage[utf8]{inputenc}

\usepackage{microtype}

\newcommand{\cutabstractup}{\vspace*{-0.1in}}

\newcommand{\cutsectionup}{\vspace*{-0.03in}}
\newcommand{\cutsectiondown}{\vspace*{-0.01in}}

\newcommand{\cutsubsectionup}{\vspace*{-0.04in}}
\newcommand{\cutsubsectiondown}{\vspace*{-0.05in}}

\newcommand{\cutparagraphup}{\vspace*{-0.03in}}

\usepackage{graphicx}
\usepackage{amsmath}
\usepackage{booktabs}
\usepackage{algpseudocode}
\usepackage{multirow}
\usepackage{enumitem}
\usepackage{cleveref}
\usepackage{tikz}
\usepackage{pgfplots}
\usepackage{mathtools}
\usepackage{csvsimple}

\usepackage{algorithm}
\usepackage{algpseudocode}
\usepackage[font=small]{caption}

\newcommand{\la}[1]{}
\usepackage{soul}

\DeclareMathOperator*{\argmax}{argmax}

\usepackage{arydshln}

\makeatletter
\def\adl@drawiv#1#2#3{%
        \hskip.5\tabcolsep
        \xleaders#3{#2.5\@tempdimb #1{1}#2.5\@tempdimb}%
                #2\z@ plus1fil minus1fil\relax
        \hskip.5\tabcolsep}
\newcommand{\cdashlinelr}[1]{%
  \noalign{\vskip\aboverulesep
           \global\let\@dashdrawstore\adl@draw
           \global\let\adl@draw\adl@drawiv}
  \cdashline{#1}
  \noalign{\global\let\adl@draw\@dashdrawstore
           \vskip\belowrulesep}}
\makeatother

\title{Few-shot Subgoal Planning with Language Models}

\author{Lajanugen Logeswaran\thanks{$\,\,\,$Correspondence to \tt{llajan@lgresearch.ai}}\hspace{3pt}, Yao Fu$^{\ddagger}$, Moontae Lee$^{*\dagger}$, Honglak Lee$^{*\ddagger}$ \\
LG AI Research$^*$, University of Illinois at Chicago$^\dagger$, University of Michigan$^\ddagger$
}

\begin{document}
\maketitle
\begin{abstract}
\cutabstractup
  Pre-trained large language models have shown successful progress in many language understanding benchmarks. This work explores the capability of these models to predict actionable plans in real-world environments. Given a text instruction, we show that language priors encoded in pre-trained language models allow us to infer fine-grained subgoal sequences. In contrast to recent methods which make strong assumptions about subgoal supervision, our experiments show that language models can infer detailed subgoal sequences from few training sequences without any fine-tuning. We further propose a simple strategy to re-rank language model predictions based on interaction and feedback from the environment. Combined with pre-trained navigation and visual reasoning components, our approach demonstrates competitive performance on subgoal prediction and task completion in the ALFRED benchmark compared to prior methods that assume more subgoal supervision.

\end{abstract}

\cutsectionup
\section{Introduction}
\cutsectiondown

Developing autonomous agents that can complete specific tasks given goal descriptions embodies human-level intelligence. 
Successful agents in this setting require multiple reasoning capabilities including natural language understanding, visual reasoning, and acting over long temporal horizons. 
Training black-box models that map instructions and observations to suitable actions has proven to be difficult due to challenges in interpreting and reasoning with multimodal information, especially in the absence of strong supervision.
Thus, generalization in this setting demands effective strategies for planning, exploration, and incorporating feedback from the environment.

\begin{figure}[!t]
\centering
\includegraphics[width=0.49\textwidth]{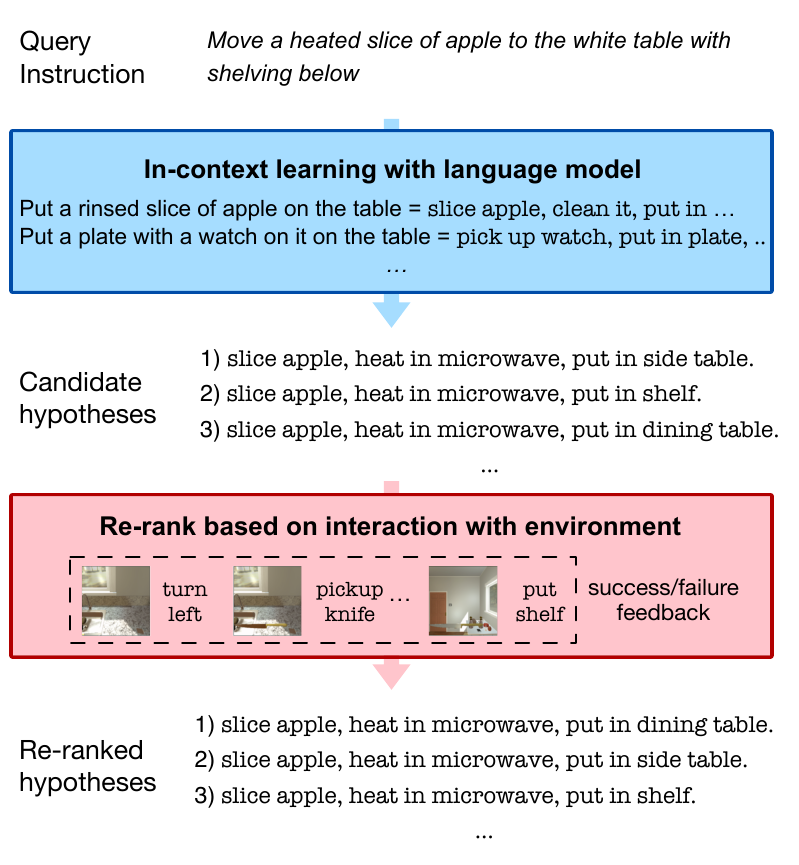}
\vspace{-0.3in}
\caption{
Overview of subgoal prediction approach.
(a) A pre-trained language model prompted with a sequence of training instances, i.e., (instruction, subgoal sequence) pairs, and a query instruction predicts top-k hypotheses using beam search.
(b) These predictions are then re-ranked by incorporating information about the environment. %
}
\label{fig:illustration}
\vspace{-0.2in}
\end{figure}

Generalization in human agents, on the other hand, stems from our ability to naturally brainstorm abstract subgoals, better calibrating executable actions and their sequences. 
Planning at the level of subgoals instead of low-level actions allows us to better adapt to unfamiliar settings.
We posit that language supervision can help realize such planning capabilities effectively in artificial agents. 
First, text is a natural API for interacting with intelligent agents that act in the real world to complete tasks.
Knowledge available in the form of text corpora, descriptions and instructions can be exploited to build better agents \citep{branavan2012learning,zhong2019rtfm}.
Second, strong language priors are useful to reason about causal sequences of events \citep{li-etal-2021-implicit}.
Language priors can further inform about object affordances (e.g. an apple is sliceable, whereas a table is not) and other contextual knowledge (e.g. a slicing task is more likely to be performed in a kitchen than a bathroom) \citep{chen2020enabling}.
Recent advances have demonstrated that large language models are able to capture such priors, as evidenced by their strong capabilities in language understanding and beyond \citep{devlin-etal-2019-bert,radford2019language,brown2020language,bommasani2021opportunities}.
This leads to the natural question of whether priors learned by language models can help reason about subgoals.

We study the ability of language models to reason about plans composed of a sequence of intermediate goals for completing basic object manipulation tasks in a household environment specified using text instructions.
In particular, we use the \textit{in-context learning} ability \citep{brown2020language} of large pre-trained language models to reason about subgoals.
In contrast to prior methods that fine-tune language models to predict subgoals/actions \citep{jansen-2020-visually,yao-etal-2020-keep}, we show that they can predict subgoal sequences effectively without any fine-tuning.
We teach the model how instructions translate into subgoal sequences by constructing a prompt using few examples.
Given the prompt and a query instruction the language model predicts likely subgoal sequences (see \Cref{fig:illustration} for an illustration). 

While language models are capable of generating strong hypotheses, we observe that these predictions may not be directly usable by agents acting in real environments. 
First, they suffer from calibration issues: Language models have a tendency to repeat content from the prompt \citep{zhao2021calibrate}.
We show that mutual-information inspired metrics help mitigate calibration issues and lead to better ranking of model generated hypotheses.

Second, real-world agents have to update their beliefs and predictions based on interaction and feedback from the environment.
Without such feedback we cannot expect the predicted plan to be executable in the environment.
We execute plans proposed by the language model in the environment using a pre-trained low-level policy and collect feedback about task success/failure.
We use this feedback as a learning signal to train a ranking model that re-ranks language model predictions. 
In contrast to prior methods that rely on strong subgoal supervision and task level expert trajectories, we show that combining subgoal predictions with a pre-trained subgoal execution policy leads to a strong embodied agent baseline.

We make the following contributions in this work. We show that
\begin{itemize}[leftmargin=*,topsep=0pt]
\setlength\itemsep{-2pt}
\item Large language models can predict subgoals from text instructions with very little supervision using in-context learning.
\item Incorporating a small amount of feedback from interaction with the environment such as agent state and task success/failure outcome improves language model predictions.
\item Combining predicted subgoals with a pre-trained low-level policy for navigation and visual reasoning leads to a simple modular agent policy that performs well on an embodied learning setting.
\end{itemize}

\cutsectionup
\section{Related work}
\cutsectiondown

\cutparagraphup
\paragraph{Language models for planning and interaction}
The use of language models for planning and action prediction has been explored in prior work.
\citet{jansen-2020-visually} fine-tuned a language model to predict subgoal sequences for text instructions from the ALFRED benchmark.
\citet{micheli-fleuret-2021-language} take a similar approach, but show that imitation learning with few instances combined with reinforcement learning produces models that work well on the ALFWorld benchmark \citep{shridhar2020alfworld}.
\citet{yao-etal-2020-keep} demonstrate a similar approach for interactive fiction games \citep{hausknecht2020interactive}.
In contrast to these prior methods, our approach does not assume strong supervision and we demonstrate generalization with limited training examples. 
Furthermore, in order to exploit the generalization capabilities of large language models, we do not fine-tune these models and instead use their in-context learning ability.
Finally, our approach allows us to build policies that inherit the strong generalization capabilities of these large pre-trained models such as compositional generalization.

\la{Contemporary work \cite{huang2022language}, \cite{ahn2022can}}

\cutparagraphup
\paragraph{Large language models and few-shot learning} 
\citet{brown2020language} showed that pre-trained large language models have few-shot learning capabilities.
Given a few examples $\{(x_i, y_i = f(x_i))\}$ that define a task $f$ such as classification or translation and a query instance $x^q$, \textit{prompting} a language model with a string such as "$x_1 = y_1; x_2 = y_2; ... ; x_n = y_n; x^q =$" leads to meaningful completions by the language model $y^q \approx f(x^q)$.
This few-shot learning capability of language models has since then been studied and improved upon with approaches like prefix engineering \cite{schick-schutze-2021-just}, prompt tuning \citep{li-liang-2021-prefix}, model calibration \cite{zhao2021calibrate} and other methods \cite{min2021noisy}.
We adopt a similar approach for few-shot subgoal inference.
We assume that subgoal supervision is available for a small number of training tasks and use the language model to infer subgoals for unseen tasks.

\cutparagraphup
\paragraph{Instruction following}

\la{This may have to be more detailed}
There is rich literature on agents that follow language instructions (\citet{branavan-etal-2009-reinforcement,mei2016listen,fried-etal-2018-unified,suhr-etal-2019-executing} \textit{inter alia}\footnote{See \citet{luketina2019survey} for a comprehensive survey}).
Recent developments in simulated environments and benchmarks with human annotated instructions have driven progress in embodied agents that learn from text instructions \citep{shridhar2020alfred,kolve2017ai2}. %
Successful agents in these settings require multiple reasoning capabilities including language understanding, visual reasoning and learning to act over long time-horizons.
Recent embodied learning literature exploit subgoal supervision, pre-trained visual reasoning components and pre-trained transformer models to do well on the task \citep{singh2020moca,suglia2021embodied,zhang-chai-2021-hierarchical,corona-etal-2021-modular,blukis2022persistent}. 
Unlike these methods, we do not assume access to strong subgoal supervision or task level expert supervision.
We combine language model predictions with pre-trained low-level navigation and interaction policies to obtain a competitive agent policy.

\cutparagraphup
\paragraph{Few-shot semantic parsing} %
Subgoal inference from text instructions can be considered a semantic parsing problem where the subgoal sequences serves as a formal representation of text.
\citet{shin-etal-2021-constrained} show that few-shot semantic parsers can be derived from language models and demonstrate their applicability on text-to-SQL \citep{finegan-dollak-etal-2018-improving} and SCAN \citep{lake2018generalization} benchmarks.
\citet{furrer2020compositional} and \citet{herzig2021unlocking} further study the compositional generalization ability of such semantic parsers.
In our work we make use of ideas introduced in these works such as dynamic prompt creation, constrained decoding and intermediate representations.

\cutsectionup
\section{Approach}
\cutsectiondown

We first consider subgoal inference as a semantic parsing problem where a text instruction needs to be translated to a sequence of subgoals and propose an approach to few-shot subgoal inference based on pre-trained language models in \Cref{sec:subgoalinference}.
We extend this setting to an agent acting in a simulated environment which can execute these subgoals, observe feedback, and improve upon language model predictions for more accurate subgoal inference in \Cref{sec:policy}.

\subsection{Few-shot subgoal inference}
\label{sec:subgoalinference}

\cutparagraphup
\paragraph{Subgoals}
We are interested in a particular sub-class of instruction following problems which involve performing a sequence of object interactions in an embodied environment.
Each object interaction requires navigating to a particular object and performing an action on it.
A task is considered successfully completed if the state of objects in the end satisfy a set of task-specific constraints (for instance, objects that need to be sliced/warmed/cooled/cleaned have the appropriate state change).
It is thus natural to define a subgoal as one or more sequence of object interactions.
A subgoal $g$ is specified as $g = (b, o) \in \mathcal{B} \times \mathcal{O}$ where $b \in \mathcal{B} = $ \{Pickup, Clean, Heat, ..\} is one of a pre-defined set of abstract actions and $o \in \mathcal{O} = $\{Apple, Microwave, DeskLamp, Ottoman, ..\} is an object category.

\cutparagraphup
\paragraph{Subgoal inference problem}
Given a text instruction $\tau$, subgoal inference seeks to predict a sequence of subgoals $\tau \mapsto g = (g^{(1)}, ..., g^{(n)})$.
To perform in-context learning with a language model, we consider a representation $v(g)$ of $g$ that looks like natural text, where $v$ is a pre-defined invertible mapping.
Such representations have been referred to in the literature as verbalizers \citep{min2021noisy}, intermediate representations \citep{herzig2021unlocking} and canonical representations \citep{shin-etal-2021-constrained}, where the purpose is to represent the output in a format the language model understands.  
In a slight abuse of notation, we will use $g$ to refer to either a subgoal sequence or it's textual representation $v(g)$ depending on the context.

\begin{figure*}[!ht]
\centering
\includegraphics[width=0.98\textwidth]{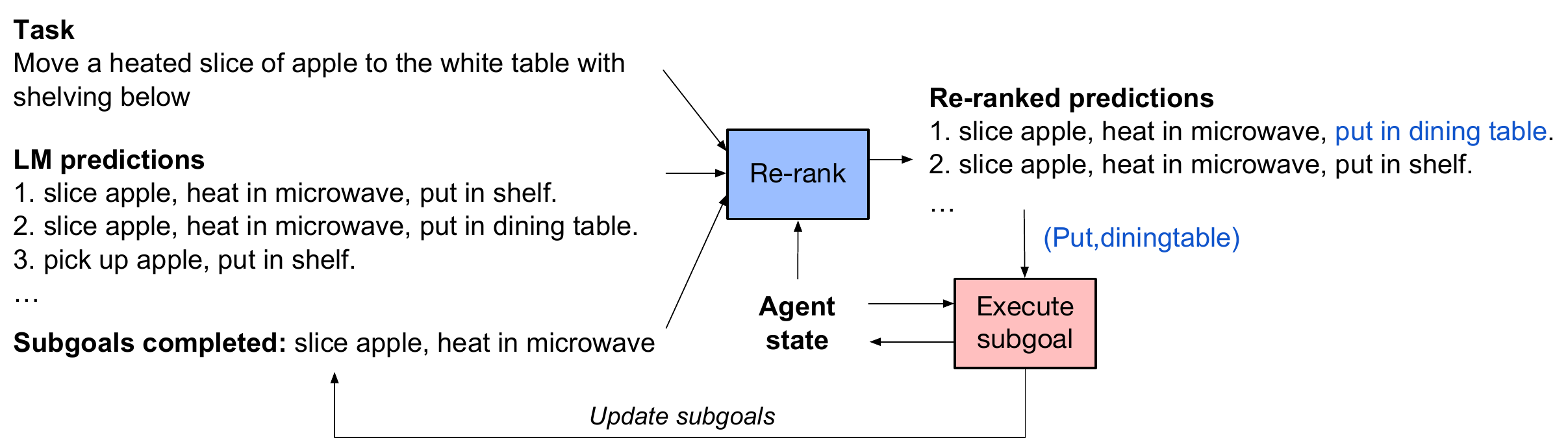}
\caption{
Re-ranking language model predictions with interaction and feedback from the environment.
Given the task, language model predictions, completed subgoals and the agent state we come up with a ranked list of subgoal sequences. 
The agent then executes the next subgoal from the highest ranked plan. %
The completed subgoal is added to the partial plan and the process continues until stop subgoal is encountered.
During training the agent receives a positive reward if the task is successfully completed, which we use as supervision to train the ranking model.
}
\label{fig:overview}
\vspace*{-0.15in}
\end{figure*}

\cutparagraphup
\paragraph{Generating subgoals}
We assume that a small amount of training data $\{(\tau_1, g_1), \cdots, (\tau_n, g_n)\}$ is given. %
The language model is prompted with a comma separated concatenation of the training examples, each in the format "$\tau_i = g_i$", followed by a query $\tau$, formatted as "$\tau = $".
We assume that the probability of a hypothesis $h$ (i.e., text representaton of a subgoal sequence) can be modeled as in \Cref{eq:prob}, where $h_i$ is the $i^\text{th}$ token of $h$ and the token probabilities are derived from the language model.
\begin{equation}
p(h|\tau) = \prod_i p_\text{LM}(h_i|h_{<i}, \tau, \{\tau_j,  g_j\}_{j=1}^n)
\label{eq:prob}
\end{equation}
We use beam search to identify the top-k hypotheses according to $p(h|\tau)$.
Generated hypotheses are constrained to be valid representations of subgoal sequences by considering only tokens which lead to a valid partial prefix of a subgoal sequence at each step of beam search.

\cutparagraphup
\paragraph{Re-ranking predictions}
Recent studies have found that language models have popularity and recency biases: the tendency to repeat content mentioned in the prompt, especially content appearing later in the prompt \citep{zhao2021calibrate}.
They considered a simple approach to mitigate such biases in model predictions for classification tasks by comparing the likelihood of an output label with and without the query.
In contrast to this `direct model' which models the probability of a label given the input $p(y|x)$, \citet{min2021noisy} showed that a `channel model' which models $p(x|y)$ leads to better, more stable models.

Inspired by these observations, we propose to use $p(\tau|h)$ to score hypotheses in addition to $p(h|\tau)$. %
Mutual Information based ranking metrics are a natural candidate and they have been explored in the text generation literature \citep{li-etal-2016-diversity,li2016mutual}.
We generate multiple hypotheses from the model using $p(h|\tau)$ and the generated hypotheses are re-scored using the weighted mutual information metric $(1 - \lambda) \text{log }p(h|\tau) + \lambda \text{log }p(\tau|h)$ where $\lambda$ is a hyperparameter\footnote{\Cref{sec:MI} details the connection to Mutual Information.}.
To compute $p(\tau|h)$, we again use the language model prompted with "$g_1=\tau_1,...,g_n=\tau_n,h=$" as the query and compute the conditional probability of $\tau$.
We expect this paradigm of generating a set of strong hypotheses, followed by accurate re-ranking is more generally applicable to other few-shot language understanding problems.

\cutsubsectionup
\subsection{Agent policy and incorporating environment feedback}
\label{sec:policy}
We next consider building an agent that acts in a visual environment to complete tasks given text instructions. 
While \Cref{sec:subgoalinference} treated the language model as a knowledge extraction system, in the real world plans need to be updated based on interaction and feedback from the environment.
We thus propose a method to improve language model predictions based on environment interaction.
Since our goal is to learn the planning component of the agent, we assume a pre-trained low-level policy is provided and optimize over the space of plans.
Jointly learning both components is beyond the scope of this work and left as future work.

We assume that a representation of the agent state $s$ is available.
The state representation captures information about the environment (e.g. objects present and their locations) estimated based on the agent's visual observations.
As the agent explores the environment and collects new observations the state representation is updated.
Assuming that a low-level policy $\pi_L$ pre-trained to execute a given subgoal is provided, our goal is to train a high-level policy $\pi_H$ which proposes the subgoals to be executed by the low-level policy.
More formally, the high-level policy models $\pi_H(g^{(t)} |\tau, s_t, g^{(<t)})$ where $\tau$ is a text instruction, $s_t$ is the state representation and $g^{(<t)}$ is the sequence of subgoals completed so far at high-level time-step $t$\footnote{Alternatively, this can be framed as a POMDP in a hierarchical reinforcement learning setting.}.

While the language model can generate compelling subgoal hypotheses, it doesn't take into account information about the environment.
For instance, knowledge about the type of room the agent is in (kitchen, bathroom, etc.) and the objects present in it are useful to infer the kind of tasks and subgoals that can be performed.
We propose to re-rank hypotheses generated by the language model based on information from the environment to construct $\pi_H$.
The plans generated by the language model are executed in the environment using $\pi_L$.
The success/failure outcomes of these plan executions are used to construct a labeled dataset of instructions $\tau$, plans $g$ and agent state $s$.
A supervised ranking model $f(g, \tau, s; \theta)$ is trained using this data to re-rank the language model predictions.
We represent the ranking model as $f(g, \tau, s; \theta) = \theta^T \text{concat}(f^\text{state}(s), f^\text{text}(\tau, g))$ where $f^\text{state}(s)$ is a state embedding, $f^\text{text}(\tau, g)$ is a joint encoding of $\tau$ and $g$ produced by a text encoder and $\theta$ is a parameter vector.
Although a text embedding can be derived from the language model, we use a BERT encoder in favor of obtaining a smaller dimensional representation ($f^\text{text} = \text{BERT}_\text{CLS}$). 
See \Cref{app:rankingarch} for more details. 

During inference, an instruction $\tau$ is given, and we use the procedure in \Cref{sec:subgoalinference} to generate top-k hypotheses. 
At each step, hypotheses inconsistent with the sequence of subgoals executed so far are pruned and the remaining hypotheses are re-ranked based on the current agent state using $f$.
The agent attempts the next subgoal proposed by the top hypothesis.
The process ends when the stop subgoal is predicted. 
See \Cref{fig:overview} for an illustration and \Cref{app:ranking} for more details about the training and inference algorithms.

\cutsectionup
\section{Experiments}

\paragraph{Data}
We use data from the ALFRED benchmark proposed by \citet{shridhar2020alfred} in our experiments. 
The ALFRED task requires an agent to execute instructions specified in text to accomplish basic tasks in an embodied environment. 
A given task is described using a high-level language directive as well as low-level step-by-step instructions (We only use the high-level description).
The dataset consists of 7 task types (and 11 fine-grained types), and has more than 20k natural language task descriptions collected from human annotators.
In addition, expert demonstrations computed by a planner are also made available.
Tasks require acting over many time-steps, with an average of 50 actions, and the longest tasks require 100+ steps.

The ground truth subgoal sequences in the dataset consist of both navigation subgoals and object interaction subgoals.
We discard the navigation subgoals and only retain the interaction subgoals for the following reasons.
First, the interaction subgoals are sufficient for an agent to successfully complete the task.
Second, predicting navigation subgoals from the text instruction alone may not always be possible as they often depend on the scene layout.

\cutparagraphup
\paragraph{Subgoal representation}
A subgoal $g^s$ is specified as $g^s = (b, o) \in \mathcal{B} \times \mathcal{O}$ where $|\mathcal{B}| = 7$ and $|\mathcal{O}| = 80$.
We define a textual representation $v(b)$ of each action type (e.g. $v$(Pickup) $=$ `pick up', $v$(Heat) $=$ `heat in').
The object types $o$ are identified by a text string $v(o)$ in the dataset and we directly use them as the text representation with minimal pre-processing (e.g. $v$(apple) $=$ `apple', $v$(desklamp) $=$ `desk lamp'). 
The subgoal is represented as $v(g^s) = $`$v(b)$ $v(o)$' (e.g. $v$((Pickup, apple)) $=$ `pick up apple').
A subgoal sequence $g = (g^{(1)}, ..., g^{(n)})$ is represented as $v(g) = $`$v(g^{(1)})$, ..., $v(g^{(n)})$.'.
Text representations of all subgoals are given in \Cref{app:subgoalrep}.
Note that there are many plausible choices for the representation $v$ and a different set of choices can lead to different results.

\cutparagraphup
\paragraph{Metrics}
We use \textit{top-k recall} to evaluate the ability of language models to generate plans from instructions by comparing against ground truth plans.
In addition, we also evaluate the performance of an agent acting in the AI2-Thor \citep{kolve2017ai2} simulator to complete tasks using \textit{task success rate} (the percentage of tasks successfully completed).

\begin{table*}[!t]
\def\arraystretch{1.1}
\small
\centering
\begin{tabular}{l c c c c c c c}
\toprule
\multicolumn{1}{c}{\multirow{3}{*}{Top-$k$ recall}} & \multicolumn{1}{c}{\multirow{3}{*}{Ranking criteria}} & \multicolumn{3}{c}{GPT2-XL} & \multicolumn{3}{c}{GPT-J} \\
\cmidrule[0.8pt](lr){3-5} \cmidrule[0.8pt](lr){6-8}
& & $N=11$ & $N=22$ & $N=33$ & $N=11$ & $N=22$ & $N=33$ \\
\midrule

\csvreader[
  column count=8,
  no head,
  table head=\\,
  late after line=\\,
  before line=\ifthenelse{\equal{\thecsvrow}{2}}{\\[-1.2em]\midrule}{}
]{tables/ranking.csv}{}
{\csvcoli & \csvcolii & \csvcoliii & \csvcoliv & \csvcolv & \csvcolvi & \csvcolvii & \csvcolviii}
\bottomrule
\end{tabular}
\vspace*{-0.05in}
\caption{
Re-ranking model generated hypotheses using different criteria. 
The first section shows top-10 recall of generated hypotheses (using $p(h|\tau)$).
The second section shows top-1 recall after re-ranking these hypotheses using different criteria. 
Results are shown for GPT2-XL and GPT-J models when the number of training instances $N$ is varied.
}
\vspace*{-0.05in}
\label{table:ranking}
\end{table*}

\begin{table}[!t]
\setlength{\tabcolsep}{5pt}
\small
\centering
\begin{tabular}{l c c c c}
\toprule
\multicolumn{1}{c}{\multirow{3}{*}{Task}} & \multicolumn{2}{c}{GPT2-XL} & \multicolumn{2}{c}{GPT-J} \\
\cmidrule[0.8pt](lr){2-3} \cmidrule[0.8pt](lr){4-5}
& top-1 & top-10 & top-1 & top-10 \\
\midrule

\csvreader[
  column count=5,
  no head,
  table head=\\,
  late after line=\\,
  before line=\ifthenelse{\equal{\thecsvrow}{8}}{\\[-1em]\midrule}{}
]{tables/main_results.csv}{}
{\csvcoli & \csvcolii & \csvcoliii & \csvcoliv & \csvcolv}
\bottomrule
\end{tabular}
\vspace*{-0.05in}
\caption{Top-k recall for subgoal sequences predicted by GPT2-XL and GPT-J models categorized by task type.
}
\vspace*{-0.2in}
\label{table:subgoalinference}
\end{table}

\cutsubsectionup
\subsection{Few-shot subgoal inference}
\label{sec:expsubgoalinference}
\cutsubsectiondown

We construct a training set of $N = 22$ instances by randomly choosing two instances per fine-grained task type.
The language model is prompted with a concatenation of these training examples and the query instance. 
We perform constrained beam search decoding with a beam size of 10 to generate subgoal sequences. 
At each step of beam search, only tokens which lead to a valid partial prefix of a subgoal sequence are considered.
All model generated hypotheses thus correspond to valid subgoal sequences.
We evaluate models on the valid-seen split of the dataset which has 800 instances.

\Cref{table:subgoalinference} shows subgoal inference results categorized by task type. 
We use publicly available pre-trained transformer language models GPT2-XL \citep{radford2019language} and GPT-J \citep{gpt-j} via the HuggingFace library \citep{wolf-etal-2020-transformers}, which respectively have 1.5B and 6B parameters, in our experiments.
The first six of the seven task types have two object arguments each.
The pick place movable task type has three object arguments and hence a lower recall than the other task types.
The top-10 recall of GPT2-XL and GPT-J are respectively 59\% and 71\%, which shows that large language models have strong ability to reason about plans from few training examples.

\cutparagraphup
\paragraph{Re-ranking hypotheses}
The top-k recall performance reported in \Cref{table:subgoalinference} is based on log $p(h|\tau)$.
We confirmed that the biases reported in the literature such as predicting content from the prompt are present in model predictions \citep{zhao2021calibrate}.
Consider the query example \textit{Place a martini glass with a fork on it on the table}.
The following two are among the top generated hypotheses:

\noindent a) pick up fork, put in cup, pick up cup, put in sink.

\noindent b) pick up fork, put in cup, pick up cup, put in table.

When the prompt contains training examples that mention `sink', the model assigns the following log $p(h|\tau)$ to these hypotheses: a) -2.4 and b) -4.3.
However, when all training instances in the prompt involving `sink' are removed, the log probabilities now become,
a) -13.7 and b) -9.1
The incorrect hypotheses involving `sink' is now ranked below the correct hypothesis involving table.
While language models can retrieve strong hypotheses as indicated by the high top-10 recall, this observation shows that the ranking of these hypotheses, as determined by $p(h|\tau)$, may not be accurate. 
We thus consider mutual information based ranking approaches.
\Cref{table:ranking} shows top-1 recall when model generated hypotheses are ranked according to different criteria.
We also vary the number of training examples $N$ by randomly choosing respectively 1, 2 and 3 instances per fine-grained task type.

We first observe that $p(\tau|h)$ ranks hypotheses better than $p(h|\tau)$ with very limited supervision ($N=11$).
However, it is often worse when more supervision is available.
In contrast, combining the two log probabilities with $\lambda = \frac{1}{2}$ yields consistently better performance across models and number of training examples.
This shows that generating a large number of hypotheses with a language model, followed by more accurate re-ranking using Mutual Information inspired metrics can be an effective paradigm for few-shot generation tasks with in-context learning.

\cutparagraphup
\paragraph{Comparison with prior work}
We compare our prediction performance against prior work in \Cref{table:subgoalcompare}.
\citet{jansen-2020-visually} fine-tunes a GPT2-Medium model (325M parameters) to predict subgoals from instructions
and report prediction results\footnote{https://github.com/cognitiveailab/alfred-gpt2} when the model is trained on varying amounts of training data: 1\%, 10\%, 25\%, 100\% of the training set, which has 7793 instances.
We ignore the navigation subgoals in this evaluation and only compare the sequence of object interactions. %
We report prediction performance of GPT-J using our approach on the same test set.
The results show that large language models encode useful knowledge that can help plan from instructions effectively when supervision is limited.
However, fine-tuning can be effective when more supervision is available due to the fixed context length limitation of in-context learning.
See \Cref{sec:ablations} for ablations and more discussion about fine-tuning.

\begin{table}[!t]
\small
\centering
\begin{tabular}{l c c}
\toprule
Model & Top-1 & Training\\
& recall & instances\\
\midrule
\multirow{4}{*}{
\shortstack[l]{\citet{jansen-2020-visually} \\ (Fine-tuned GPT2-Medium)}
} 
& \phantom{0}5.07 & \phantom{00}77 \\
& 41.92 & \phantom{0}779 \\
& 53.80 & 1948 \\
& 61.00 & 7793 \\
\cdashlinelr{1-3}
\multirow{3}{*}{
\shortstack[l]{Ours \\ (In-context GPT-J)}
}
& 44.02 & \phantom{00}11 \\
& 47.80 & \phantom{00}22 \\
& 49.63 & \phantom{00}33 \\
\bottomrule
\end{tabular}
\vspace*{-0.05in}
\caption{
Comparison against subgoal prediction performance of \citet{jansen-2020-visually}.
}
\vspace*{-0.1in}
\label{table:subgoalcompare}
\end{table}

\cutparagraphup
\paragraph{Prediction errors}
We examine prediction errors in identifying task type and object type.
Key sources of model errors include annotation issues and ambiguity in object types.
\Cref{table:confusion} shows the object types that have the least prediction accuracy, along with the object categories the model is confused about. %
Annotations can fail to correctly identify the target object - identifying a \emph{butter knife} as a \emph{knife} or a \emph{pepper shaker} as \emph{salt shaker}.
Ambiguity can also arise from identifying an object with different names, depending on the context.
For instance, depending on the scene layout, the argument for a \textit{look at object in light} task can be a floor lamp or a desk lamp.
Unless the type of lamp is identified precisely in the instruction, it is not possible to correctly predict the type of lamp. %
Correctly identifying these objects requires feedback from interaction with the environment.

The experiments so far evaluate the ability of a language model to retrieve ground truth subgoal sequences.
Next we examine embodied agents that make use of these predictions and collect more supervision in order to improve subgoal predictions.

\subsection{Agent policy and incorporating environment feedback}

We now use subgoal predictions to construct an agent policy that acts in a simulated environment to complete tasks.
The agent state representation and pre-trained low-level subgoal policy are borrowed from the HLSM model proposed in \citet{blukis2022persistent}.
HLSM represents the agent state as a spatial persistent voxel representation of the room which models the location and category of objects present.
The representation is constructed using modules that estimate segmentation and depth maps and other visual reasoning components and is updated as the agent gathers new observations. 
We use pre-trained models made available by the authors\footnote{https://hlsm-alfred.github.io} for state estimation and the low-level policy in our experiments.

We combine subgoal predictions with the pre-trained HLSM low-level policy and evaluate the overall agent policy on the ALFRED task in \Cref{table:alfred}.
Unlike the results reported in \Cref{sec:expsubgoalinference} which were based on the static dataset, these results are based on subgoals executed against the AI2-Thor simulator.
In addition to task success rate, we also report the percentage of goal conditions satisfied, which rewards the model for partial task completions.

\begin{table}[!t]
\small
\centering
\begin{tabular}{l l}
\toprule
Obj category & Confusion categories \\
\midrule
wateringcan
& pencil, kettle \\
glassbottle 
& vase \\
cart
& shelf, sidetable, microwave, cart, fridge \\
butterknife
& knife, butterknife \\ %
floorlamp
& desklamp, floorlamp \\ %
vase
& pencil, vase, bowl, winebottle, pot \\
ladle
& spoon, ladle \\ %
pot
& pot, pan \\ %
soapbottle
& soapbottle, winebottle \\ %
\bottomrule
\end{tabular}
\vspace*{-0.05in}
\caption{
Object categories the model makes most errors on and the top object categories it confuses with.
}
\vspace*{-0.1in}
\label{table:confusion}
\end{table}

We compare against the following baselines on the ALFRED task.
Seq2seq \citep{shridhar2020alfred} is a simple sequence-to-sequence baseline trained to map text instructions to low-level actions.
MOCA \citep{singh2020moca} improves on Seq2seq using subgoal supervision and pre-trained visual reasoning components.
Recent work such as HLSM \citep{blukis2022persistent} and FiLM \citep{min2021film} build and use spatial semantic state representations and achieve stronger performance on the task.
Note that, unlike these prior methods (MOCA, HLSM, FiLM) that rely on full subgoal supervision (20k instances), our approach is based on a small amount of subgoal supervision and additional supervision collected using active interaction with the environment.
In addition, our approach does not require task-level expert trajectories and only assumes that a subgoal execution policy is provided.

Using the top language model prediction as is without using any information from the environment leads to 20\% success rate.
Next, we collect plan execution feedback for 1000 text instructions to train the ranking model described in \Cref{sec:policy}.
Re-ranking language model predictions using the trained ranking model improves the performance to 24\%, which shows the importance of incorporating feedback from environment interaction.
In comparison, the HLSM model with full subgoal supervision has success rate 30\%. 
Although our predictions fall short of HLSM, they are competitive with the other baselines with subgoal supervision.
The performance upper bound estimated using oracle subgoals is 37\%, which shows the room for improvement over our predictions.
These results show that accurate subgoal inference coupled with pre-trained low-level components leads to agents that perform well in embodied environments.

\Cref{fig:illustration} shows an example where the ranking model uses environment information to identify better plans.
In this example, the instruction ambiguously specifies the receptacle as `white table with shelving'.
The language model's top two predictions for the target receptacle are `side table' and `shelf', neither of which are present in the environment. %
The agent state captures this information and helps identify `dining table' as the correct receptacle.

\begin{table}[!t]
\small
\centering
\begin{tabular}{l l c c}
\toprule
\multicolumn{2}{c}{\multirow{2}[3]{*}{Model}} & \multicolumn{2}{c}{Success rate} \\
\cmidrule[0.4pt](lr){3-4}
& & Task & Goal-Cond \\
\midrule
\multicolumn{2}{l}{Seq2seq \citep{shridhar2020alfred}}  & 3.7 & 10.0 \\
\multicolumn{2}{l}{MOCA \citep{singh2020moca}}          & 19.2 & 28.5 \\
\multicolumn{2}{l}{FiLM \citep{min2021film}}            & 24.6 & 37.2 \\
\multicolumn{2}{l}{HLSM \citep{blukis2022persistent}}   & 29.6 & 38.8 \\
\midrule
\multirow{3}{*}{\shortstack[c]{HLSM \\[-1pt] low-level \\[-1pt] policy}}
& \textit{Predicted subgoals}   & 19.8 & 31.4 \\
& \textit{Re-ranked subgoals}   & 23.9 & 35.0 \\
& Oracle subgoals               & 37.2 & 48.2 \\
\bottomrule
\end{tabular}
\caption{
Task completion and goal condition success rates of models on the ALFRED validation seen split (results are based on task executions in the AI2-Thor simulator). 
The performance of our subgoal predictions combined with the HLSM low-level policy are shown at the bottom.
We show the performance before and after re-ranking language model predictions based on agent state. 
Oracle subgoals shows the performance upper bound.
}
\vspace*{-0.1in}
\label{table:alfred}
\end{table}

\cutsectionup
\section{Ablations}
\label{sec:ablations}
\cutsectiondown

We perform a series of ablations to identify the robustness of model predictions. %
We compare the performance of in-context learning to a sequence-to-sequence model fine-tuned to translate instructions to subgoal sequences. %
In addition, we observe the effect of varying the number of training examples and choice of training examples.

\begin{figure}
\centering
\hspace{-1.5em}
\resizebox{0.37\textwidth}{!}{
\centering
\begin{tikzpicture}
\begin{axis}[
xlabel=Number of training examples,
ylabel=Recall,
legend pos=south east,
no markers,
legend style={draw=none},
xtick style={draw=none},
ytick style={draw=none},
yticklabel style={xshift=-1ex},
grid=major,
xmin=0,
xmax=45,
ymin=0,
ymax=80,
xtick={11, 22, 33, 44},
legend cell align={left},
]
\addplot+[smooth,mark=*,blue,mark size=1pt,error bars/.cd, y dir=both, y explicit,
    error bar style={line width=1pt},
    error mark options={
      rotate=90,
      blue,
      mark size=2pt,
      line width=1pt
    }
] 
coordinates {
(0, 23.29)
(11, 49.634)        +- (1.772534908, 1.772534908)
(22, 53.902)        +- (1.987981388, 1.987981388)
(33, 54.88)         +- (1.73,  1.73)
(44, 56.8)          +- (0.7, 0.7)
};
\addlegendentry{GPT-J in-context (top-1)}

\addplot+[dashed,smooth,mark=x,blue,mark size=2pt,error bars/.cd, y dir=both, y explicit,
    error bar style={line width=1pt,solid},
    error mark options={
      rotate=90,
      blue,
      mark size=2pt,
      line width=1pt
    }
] 
coordinates {
(0, 11.95)
(11, 39.512)    +- (3.054098885, 3.054098885)
(22, 45)        +- (2.196667931, 2.196667931)
(33, 46.44)     +- (2.998357884, 2.998357884)
(44, 48.172)    +- (2.755969884, 2.755969884)
};
\addlegendentry{T5-large finetune (top-1)}

\addplot+[smooth,mark=*,red,mark size=1pt,error bars/.cd, y dir=both, y explicit,
    error bar style={line width=1pt},
    error mark options={
      rotate=90,
      red,
      mark size=2pt,
      line width=1pt
    }
] 
coordinates {
(0, 31.95)
(11, 69.488)        +- (1.878515371, 1.878515371)
(22, 73.072)        +- (2.323406981, 2.323406981)
(33, 74.682)        +- (2.7, 2.7)
(44, 78.3225)       +- (1.1, 1.1)
};
\addlegendentry{GPT-J in-context (top-10)}

\addplot+[dashed,smooth,mark=x,red,mark size=2pt,error bars/.cd, y dir=both, y explicit,
    error bar style={line width=1pt,solid},
    error mark options={
      rotate=90,
      red,
      mark size=2pt,
      line width=1pt
    }
] 
coordinates {
(0, 41.1)
(11, 65.926)    +- (1.544354234, 1.544354234)
(22, 68.244)    +- (0.5309237233, 0.5309237233)
(33, 70.318)    +- (2.440362678,  2.440362678)
(44, 73.95)     +- (1.435322263 , 1.435322263)
};
\addlegendentry{T5-large finetune (top-10)}

\end{axis}
\end{tikzpicture}
}
\vspace*{-0.1in}
\caption{
Comparison between GPT-J with in-context learning and a fine-tuned T5-large model for varying number of training examples.
See text for details.
}
\label{fig:numtrain}
\vspace*{-0.1in}
\end{figure}
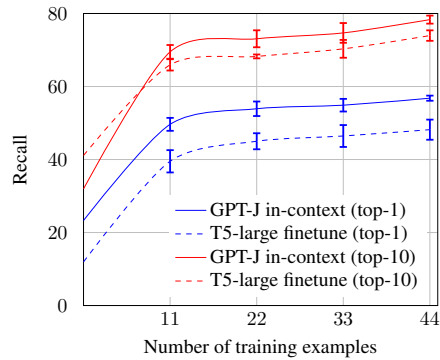

\cutparagraphup
\paragraph{Number of training examples}

\Cref{fig:numtrain} shows model recall for varying number of training examples.
For zero training examples, the prompt consists of just the query instruction and the model decodes subgoals.
Beyond 44 examples (4 examples per task type), the model prompt no longer fits the sequence length restriction (1024 tokens) of GPT models.
A steady increase in performance can be initially observed when increasing the number of training examples and the performance saturates towards the end.
In-context learning further has the limitation of not being able to accommodate a larger number of training examples due to the length restriction.
It would be interesting to explore how to make effective use of large number of training examples in future work.

\paragraph{Choice of training examples}
We also estimate performance variance by varying the random seed for choosing examples randomly from the training set and compute standard deviation based on five random seeds for each setting.
The plot shows that top-1 predictions from in-context learning have lower variance compared to fine-tuning.

\cutparagraphup
\paragraph{Comparison with fine-tuning}
In order to understand how well the in-context learning approach compares to fine-tuned models, we fine-tune a T5-large model \citep{raffel2019exploring} with 770M parameters on varying amounts of training data (this was the largest model we could fine-tune on our compute infrastructure). 
Note that this is not a head-to-head comparison between in-context learning and fine-tuning due to the difference in model size. 
Furthermore, there are other fine-tuning mechanisms such as prompt tuning and head tuning \citep{min2021noisy} which are not considered here.
However, the result suggests that in-context learning with large pre-trained models can be favorable when computational constraints do not allow full fine-tuning of large models.

These ablations show that the in-context learning ability of large language models leads to predictions that are accurate, robust and stable in the presence of a small amount of training data.

\cutsectionup
\section{Conclusion}
\cutsectiondown

This work explores the use of pre-trained language models for planning in real-world tasks.
We showed that language models have strong capability to reason about subgoal sequences given a small number of training examples.
We further demonstrated some simple mechanisms to incorporate feedback from interaction with the environment and show that this leads to more usable predictions.
Finally, we show that combining subgoal predictions with a pre-trained low-level policy yields a strong baseline for embodied agent learning.

Our ablations demonstrate that in-context learning with a small amount of subgoal demonstrations has robust generalization properties. 
However, we also point out that in-context learning has the limitation of not being able to incorporate a large number of training examples due to the fixed context length restriction.
It would further be beneficial to perform end-to-end learning with language model based subgoal prediction and a low-level policy, which would be interesting to explore in future work.

\bibliography{anthology,custom}
\bibliographystyle{acl_natbib}

\appendix

\clearpage
\newpage

\section{Mutual Information based scoring}
\label{sec:MI}

Mutual Information between random variables $X, Y$ is defined as in \Cref{eq:MI}. 
We consider a weighted Mutual Information metric as defined as in \Cref{eq:wMI} similar to \citet{li2016mutual} and introduce the hyperparameter $\lambda$. 
Identifying $Y$ that maximizes the weighted Mutual Information is equivalent to maximizing the expression in \Cref{eq:maxwMI}. 
We use this metric to rank hypotheses generated by the language model.

\begin{align}
\text{MI}(X, Y) &= \text{log }\frac{p(x, y)}{p(x) p(y)} \label{eq:MI} \\
\text{wMI}(X, Y) &= \text{log }\frac{p(x, y)}{p(x) p(y)^\lambda} \label{eq:wMI}
\end{align}

\begin{align}
&\argmax_y \text{wMI}(X, Y) \nonumber \\
&= \argmax_y \text{log }\frac{p(x, y)}{p(x) p(y)^\lambda} \nonumber \\
&= \argmax_y \text{log }\left(\frac{p(x, y)}{p(x)}\right)^{1-\lambda} \left(\frac{p(x, y)}{p(y)}\right)^\lambda \frac{1}{p(x)^\lambda} \nonumber \\
&= \argmax_y (1-\lambda) \text{log }p(y|x) + \lambda \text{log }p(x|y) \nonumber \\[-0.7em]
&\hspace{15em} - \lambda \text{log }p(x) \nonumber \\
&= \argmax_y (1-\lambda) \text{log }p(y|x) + \lambda \text{log }p(x|y)
\label{eq:maxwMI}
\end{align}

\cutsectionup
\section{Subgoal representation}
\label{app:subgoalrep}
\Cref{table:subgoalrepresentation} shows the subgoal representation we use in this work. 
\begin{table}[!h]
\captionsetup{font=normalsize}
\centering
\begin{tabular}{l l}
\toprule
Subgoal & Representation \\
\midrule
(Pickup, X) & pick up X \\
(Put, X) & put in X \\
(Heat, X) & heat in X \\
(Cool, X) & cool in X \\
(Clean, X) & clean in X \\
(Slice, X) & slice X \\
(ToggleOn, X) & turn on X \\
\bottomrule
\end{tabular}
\caption{
Subgoals and corresponding text representation. 
X represents an object argument.}
\label{table:subgoalrepresentation}
\end{table}

\newpage
\section{Ranking model: Architecture}
\label{app:rankingarch}

\paragraph{State embedding} 
HLSM represents the agent state as a semantic voxel representation $s \in [0, 1]^{X\times Y\times Z\times C}$ where the value $s(x,y,z,c)$ represents if there is an object of type $c$ at position $(x,y,z)$ of the room layout.
We pool across the spatial dimensions of the representation and project it using a linear mapping to obtain $f^\text{state}(s)$.
We use this encoding as the state embedding.

\paragraph{Instruction and subgoal sequence encoding} 
The instruction $\tau$ and a candidate subgoal sequence $g$ are jointly processed using a BERT encoder and the CLS representation is used as a representation vector $\text{BERT}_\text{CLS}(\tau, g)$.
The ranking model is represented as $f(g, \tau, s; \theta) = \theta^T \text{concat}(f^\text{state}(s), \text{BERT}_\text{CLS}(\tau, g))$ where $\theta$ is a parameter vector.

\section{Ranking Model: Training and Inference}
\label{app:ranking}

Formally, the learning problem is a MDP $(\mathcal{S}, \mathcal{G}, \mathcal{L}, \mathcal{R}, \mathcal{T})$, where 
$s_t \in \mathcal{S}$ is the agent state,
$g^{(t)} \in \mathcal{G}$ is a subgoal, %
$\tau \in \mathcal{L}$ is a text instruction, 
$\mathcal{R}(\tau, s_t)$ is a reward function that provides success/failure feedback for completing a given instruction, %
$\mathcal{T}: (s_t, g^{(t)}) \rightarrow s_{t+1}$ is a state transition function where $s_{t+1}$ is computed by a low-level policy $\pi_L$ pre-trained to execute a given subgoal $g$.
Our goal is to train a high-level policy $\pi_H(g^{(t)} |\tau, s_t, g^{(<t)})$ where $s_t$ is the agent state and $g^{(<t)}$ is the sequence of subgoals completed so far at high-level time-step $t$.

\Cref{alg:training} describes how we collect training data to train the ranking model.
\Cref{alg:inference} shows how the ranking model is used during inference.

\onecolumn

\begin{algorithm}[!ht]
\caption{Training}
\label{alg:training}
\begin{algorithmic}
\State \textbf{Given:} $\text{epochs}=100$, $\mathcal{D}^\text{ins}$ (set of instructions)
\vspace{0.3em}
\State \underline{\textit{Collect training data}}
\State $\mathcal{D} \gets \{\}$ (Initialize training set)
\For{$\tau$ in $\mathcal{D}^\text{ins}$}
  \State Generate plans and re-rank using mutual information metric $g_1, \ldots, g_k \sim p_\text{LM}(\cdot|\tau)$
  \For{$i = 1\ldots k$}
    \State Initialize agent state $s$
    \State $S\gets\{s\}$ (Record agent states)
    \For{$j = 1\ldots |g_i|$}
      \State $s \gets \mathcal{T}(s, g_i^{(j)})$ (Execute $g_i^{(j)}$ using $\pi_L$)
      \State $S \gets S\cup \{s\}$
    \EndFor
    \If {$\mathcal{R}(\tau, s) > 0$} (Task succeeded)
      \State $\mathcal{D} \gets \mathcal{D} \cup \{(g_i, \tau, s)|s \in S\}$
      \State \textbf{break}
    \EndIf
  \EndFor
\EndFor
\vspace{0.3em}
\State \underline{\textit{Train model}}
\For{$i = 1\ldots $ epochs}
  \State loss $\gets 0$
  \For{$(g, \tau, s) \in \mathcal{D}$}
    \State Generate plans $g_1,.., g_k \sim p_\text{LM}(\cdot|\tau)$
    \State loss $\gets$ loss $-$ log $\frac{\text{exp } f(g, \tau, s; \theta)}{\sum_{i=1}^k \text{exp } f(g_i, \tau, s; \theta)}$
  \EndFor
  \State $\theta \gets \text{Optimizer Update}(\theta, \nabla_\theta f)$
\EndFor
\State \Return $f$
\end{algorithmic}
\end{algorithm}

\begin{algorithm}[!h]
\caption{Inference}\label{alg:inference}
\begin{algorithmic}
\State \textbf{Given:} Instruction $\tau$, $i^\text{thresh} = 10$
\State Generate plans $g_1, \ldots, g_k \sim p_\text{LM}(\cdot|\tau)$
\State $G \gets \{g_1, \ldots, g_k\}$
\State Initialize agent state $s$
\State $i \gets 0$ (subgoal index)
\State $g \gets \argmax_{g \in G} f(g, \tau, s; \theta)$
\While{$|G| \neq 0$ and $g^{(i)} \neq $ <stop> and $i < i^\text{thresh}$}
  \State $s \gets \mathcal{T}(s, g^{(i)})$ (Execute $g^{(i)}$ using $\pi_L$)
  \State $G \gets \{h | h \in G \text{ and } h^{(i)} = g^{(i)}\}$ (Retain plans consistent with subgoals completed so far)
  \State $g \gets \argmax_{g \in G} f(g, \tau, s; \theta)$
  \State $i \gets i+1$
\EndWhile
\State \Return $g, s$
\end{algorithmic}
\end{algorithm}

\clearpage
\newpage

\section{Analysis of model errors}
\label{sec:confusion}

\begin{figure*}[!ht]
\centering
\includegraphics[angle=270,trim=102em 6em 21em 15.5em,clip,width=0.9\textwidth]{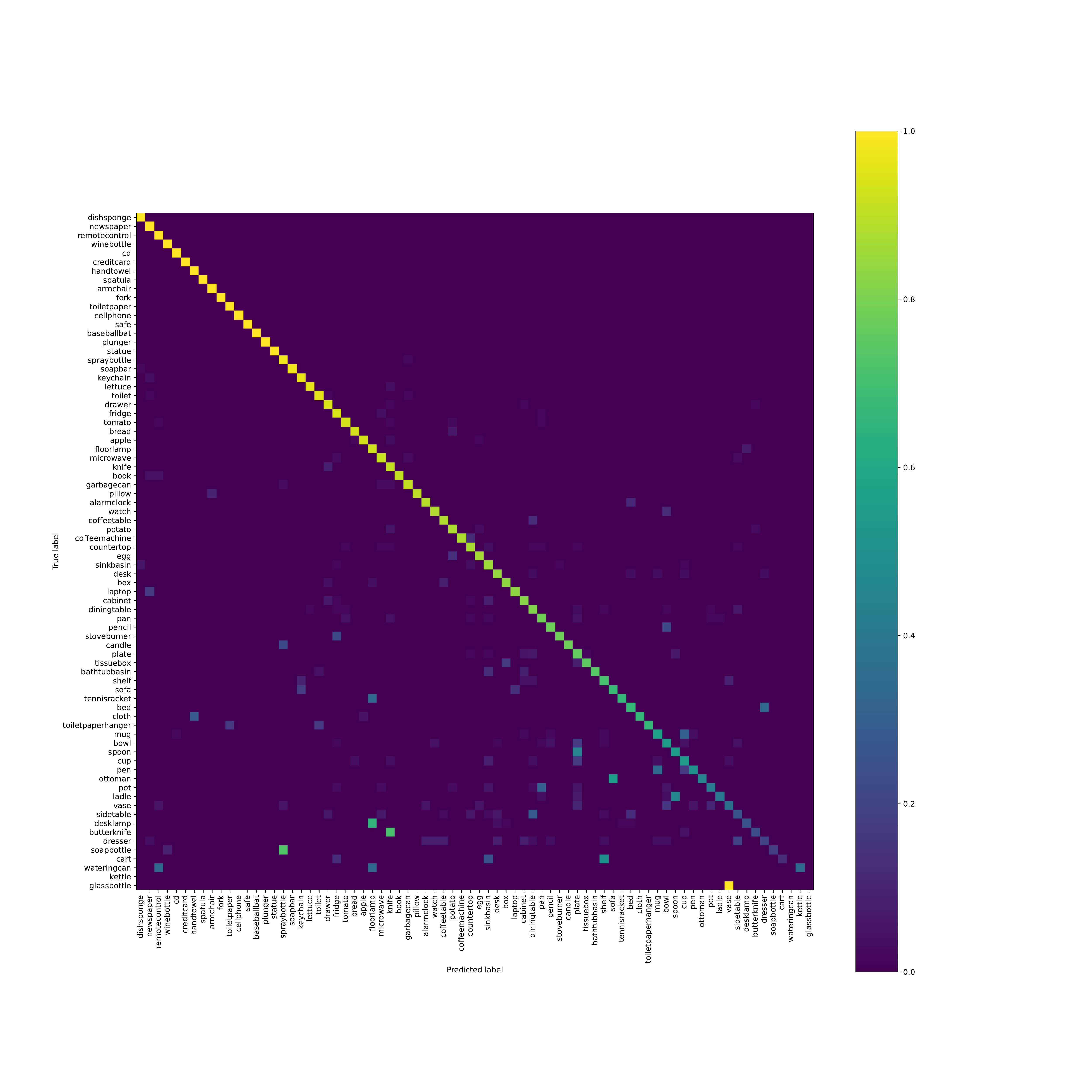}
\includegraphics[trim=6em 12em 32em 22em,clip,width=\textwidth]{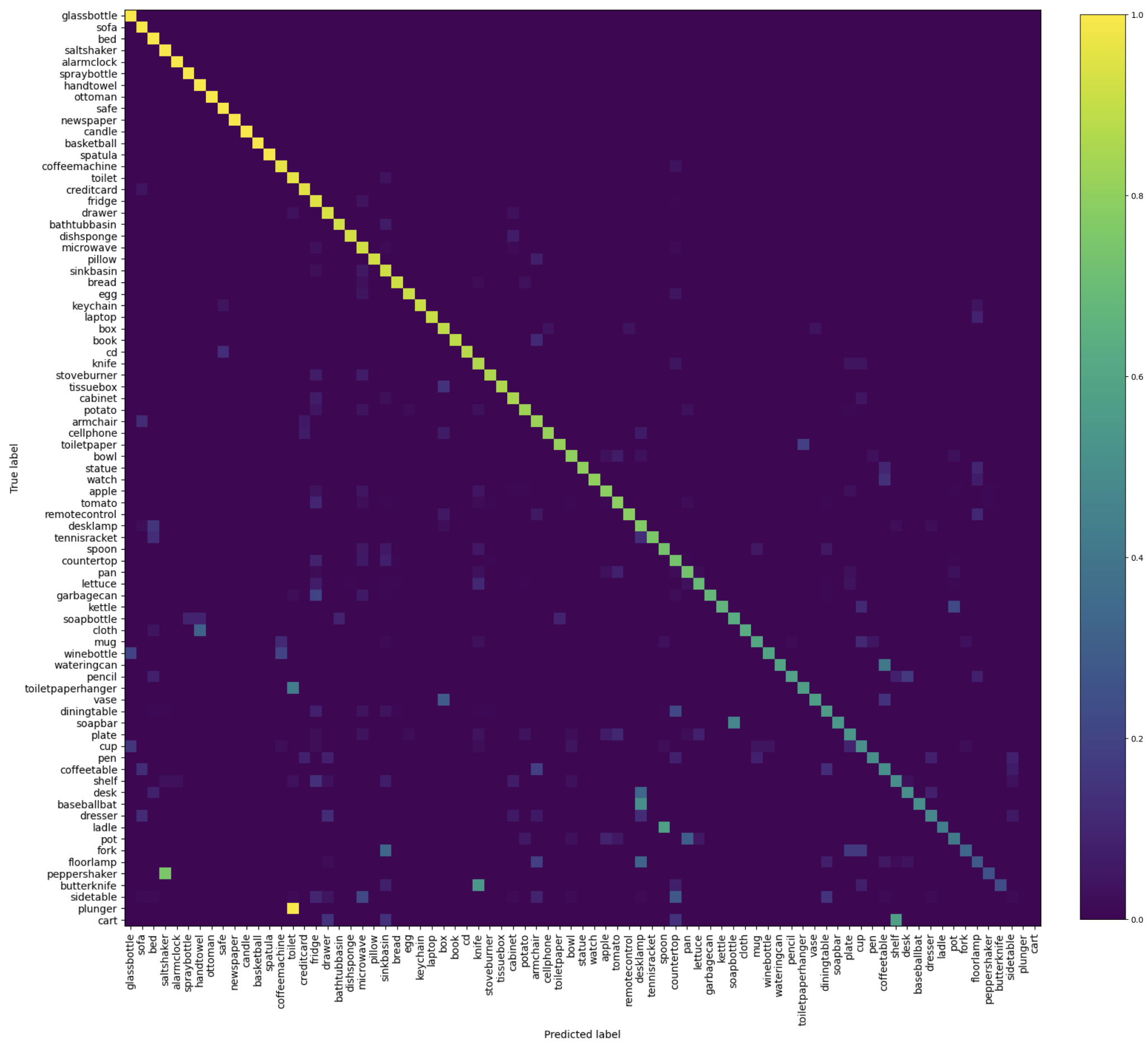}
\caption{Object type prediction confusion matrix.}
\label{figure:objectconfusion}
\end{figure*}

Figure \ref{figure:objectconfusion} shows the confusion matrix for object type prediction.
Predictions are from top-1 subgoal sequences predicted by GPT-J.

\end{document}